# SCA-Net: Spatial-Contextual Aggregation Network for Enhanced Small Building and Road Change Detection

Emad Gholibeigi, Abbas Koochari, Azadeh ZamaniFar

Department of Computer Engineering, SR.C., Islamic Azad University, Tehran, Iran.

gholibeigiem@gmail.com, koochari@gmail.com, azamanifar@gmail.com

*Abstract*— Automated change detection in remote sensing imagery is critical for urban management, environmental monitoring, and disaster assessment. While deep learning models have advanced this field, they often struggle with challenges like low sensitivity to small objects and high computational costs. This paper presents SCA-Net, an enhanced architecture built upon the Change-Agent framework for precise building and road change detection in bi-temporal images. Our model incorporates several key innovations: a novel Difference Pyramid Block for multi-scale change analysis, an Adaptive Multi-scale Processing module combining shape-aware and high-resolution enhancement blocks, and multi-level attention mechanisms (PPM and CSAGate) for joint contextual and detail processing. Furthermore, a dynamic composite loss function and a four-phase training strategy are introduced to stabilize training and accelerate convergence. Comprehensive evaluations on the LEVIR-CD and LEVIR-MCI datasets demonstrate SCA-Net's superior performance over Change-Agent and other state-of-the-art methods. Our approach achieves a significant 2.64% improvement in mean Intersection over Union (mIoU) on LEVIR-MCI and a remarkable 57.9% increase in IoU for small buildings, while reducing the training time by 61%. This work provides an efficient, accurate, and robust solution for practical change detection applications.

## I. INTRODUCTION

The rapid pace of global urbanization necessitates efficient and accurate monitoring of land surface changes. High-resolution remote sensing imagery, with its extensive spatial coverage and temporal revisit capability, has become an indispensable tool for this purpose. Within this domain, change detection (CD)—the process of identifying meaningful differences between images of the same geographical area captured at different times—is one of the most challenging and vital tasks. It finds applications in urban planning, illegal construction monitoring, disaster damage assessment, and infrastructure development tracking [1], [2].

Despite significant progress driven by deep learning, particularly Convolutional Neural Networks (CNNs) and more recently Vision Transformers (ViTs), critical challenges remain. First, low sensitivity to small objects persists, leading to high omission rates for compact buildings and narrow roads due to their limited pixel representation. Second, pseudo-changes (or false positives) caused by variations in illumination, sensor angle, and seasonal conditions can significantly degrade reliability. Third, the inherent class imbalance in CD datasets, where over 90% of pixels typically belong to the "no-change" class, heavily biases models towards the majority class. Finally, the high computational complexity and long training times of advanced architectures hinder their practical deployment and iterative development.

Recent models like Change-Agent [3], built on a Siamese SegFormer backbone, have shown promising results by integrating bi-temporal iterative interaction and fusion modules. Concurrently, novel paradigms such as Referring Change Detection (RCD) [4] and architectures based on State-Space Models (e.g., ChangeMamba [5]) have emerged, focusing on user interaction and modeling long-range dependencies. However, performance in the precise detection of small objects remains a persistent challenge.

In this paper, we propose SCA-Net (Spatial-Contextual Aggregation Network), a comprehensive enhancement of the Change-Agent architecture designed to address these limitations. Our main contributions are fourfold. First, we introduce key architectural innovations including a novel Difference Pyramid Block for robust multi-scale change analysis and an Adaptive Multi-scale Processing module. This module utilizes specialized sub-networks—a MultiScaleShapeModule for low-resolution levels and HighResEnhance blocks for high-resolution levels—to effectively handle objects of vastly different sizes. Second, we incorporate advanced attention mechanisms by integrating a Pyramid Pooling Module (PPM) [6] for aggregating global context and a Channel and Spatial Attention Gate (CSAGate) [7] for fine-grained detail refinement. This combination enables the model to jointly comprehend the overall scene context and focus on crucial local changes. Third, we design a stable and efficient training strategy centered on a dynamic composite loss function. This function strategically combines Cross-Entropy, Dice, Lovász, and Contrastive losses across a four-phase training schedule, coupled with layer-wise learning rates and coordinated data augmentation. This integrated strategy ensures stable convergence, achieves superior accuracy, and reduces total training time by 61%. Finally, we provide a comprehensive evaluation demonstrating SCA-Net's superior performance on standard benchmarks. The most significant achievement is a remarkable 57.9% relative improvement in IoU for small buildings on the LEVIR-MCI dataset, effectively addressing a fundamental challenge in the field.

## II. RELATED WORK

Traditional Change Detection Methods form the foundation of the field, encompassing techniques such as image algebra (e.g., image differencing [8]), transformation-based methods (e.g., Change Vector Analysis), and classification-based methods (e.g., Post-Classification Comparison) [9]. Although computationally efficient, these methods often lack robustness to complex radiometric and geometric variations in high-resolution imagery, making them highly susceptible to noise and false alarms.

Deep Learning-Based Methods have subsequently become the prevailing approach. Fully Convolutional Networks (FCNs) and encoder-decoder architectures like U-Net [10] became popular for dense, pixel-wise prediction. Siamese networks [11] proved particularly suitable for CD, utilizing weight-sharing branches to extract comparable features from bi-temporal images. This principle has been extended in modern models like SNUNet [12] and ChangeFormer [13], which combine Siamese encoders with sophisticated decoders. More recent works explore new paradigms. Referring Change Detection (RCD) [4] allows users to specify changes of interest via natural language prompts. Architectures like ChangeMamba [5] leverage State-Space Models for efficient long-range dependency modeling. Despite these advances, precise detection of small objects and distillation of meaningful changes from spurious ones remain significant challenges.

Multi-scale Processing and Attention Mechanisms are key concepts adopted to tackle CD challenges. Architectures like Feature Pyramid Networks (FPN) [14] are standard for handling objects of different sizes. Attention mechanisms allow models to focus on informative features. The Squeeze-and-Excitation (SE) block [15] enables channel-wise attention, while the Pyramid Pooling Module (PPM) [6] captures multi-scale spatial context. The Convolutional Block Attention Module (CBAM) [7] sequentially refines features through channel and spatial attention.

Change-Agent[3] is a significant recent architecture that integrates several advanced concepts. It employs a SegFormer backbone, a dedicated Bi-temporal Iterative Interaction (BI³) layer with Local Perception Enhancement (LPE) and Global Difference Fusion Attention (GDFA) modules, and a multi-scale decoder. While representing a robust baseline, we identify opportunities for enhancement, particularly in stabilizing its training and boosting sensitivity to small objects. SCA-Net builds directly upon this foundation, introducing targeted enhancements to its core components and strategically incorporating advanced attention and multi-scale processing concepts.

## III. PROPOSED METHOD

3.1. Overview

We name our proposed model SCA-Net (Spatial-Contextual Aggregation Network). Its overall architecture, illustrated in Fig. 1, builds upon a Siamese SegFormer-B1 backbone [16] and integrates several novel components for robust change detection. The pipeline begins with the backbone extracting multi-scale features from the input bi-temporal image pair (Image A and Image B). These features are processed through an enhanced Bi-temporal Interaction (BI³Layer) module. Subsequent adaptive processing includes the MultiScaleShapeModule for low-resolution levels and the HighResEnhance block for high-resolution levels. A novel Difference Pyramid Block (DPB) hierarchically refines change features across scales. The refined difference maps from the DPB are then concatenated with the corresponding feature maps from the adaptive processing branch and fed into the subsequent fusion modules (CBFModule and CSAGate), followed by global context aggregation via the Pyramid Pooling Module (PPM). Finally, a top-down decoder progressively upsamples the feature maps to generate the final change mask.

3.2. Siamese Backbone and Feature Extraction

We employ a Siamese SegFormer-B1 [16] as our feature extraction backbone. The network processes two bi-temporal images, denoted as $I_1$ (time T1) and $I_2$ (time T2), through a shared-weight encoder. This encoder generates four levels of multi-scale feature maps for each input image, represented as:

$$\{F_1^{(i)}, F_2^{(i)}\} \text{ for } i \in \{1, 2, 3, 4\} \quad (1)$$

where i=1 corresponds to the highest-resolution feature level (stride 1/4).

3.3. Enhanced Bi-temporal Interaction (BI³) Layer

This layer facilitates interaction between features from the two time periods, with key enhancements for improved stability and performance. The Enhanced Local Perception Enhancement (LPE) module incorporates a Squeeze-and-Excitation (SE) attention block [15]. It employs parallel convolutions with kernel sizes of 3×3, 5×1, and 1×5 to capture multi-scale local context. The outputs from these parallel pathways are concatenated and then processed by the integrated SE block, which performs dynamic channel-wise recalibration to emphasize informative features. For the Stabilized Global Difference Fusion Attention (GDFA) module, we introduce a conditional stability mechanism that reduces the internal dimensionality for layers with more than 128 channels, preventing potential numerical instability in large matrix operations [17]. Dropout layers [18] are also added to this module to mitigate overfitting.

3.4. Novel Difference Pyramid Block (DPB)

The DPB explicitly models and refines change information across scales. It takes corresponding feature pairs $\{F_1^{(i)}, F_2^{(i)}\}$ as input.

The absolute difference

$$D^{(i)} = |F_1^{(i)} - F_2^{(i)}| \quad (2)$$

is computed at each level.

Each difference map is initially refined by a 1×1 convolution. A hierarchical refinement process then proceeds from the deepest (i=4) to the shallowest level (i=1). At each level i, the upsampled and refined difference map from level i+1 is fused with the current level's refined map, followed by a 3×3 convolution for smoothing. This design ensures high-level semantic understanding of change guides precise localization at lower levels, inspired by FPN [14] but tailored for change feature propagation.

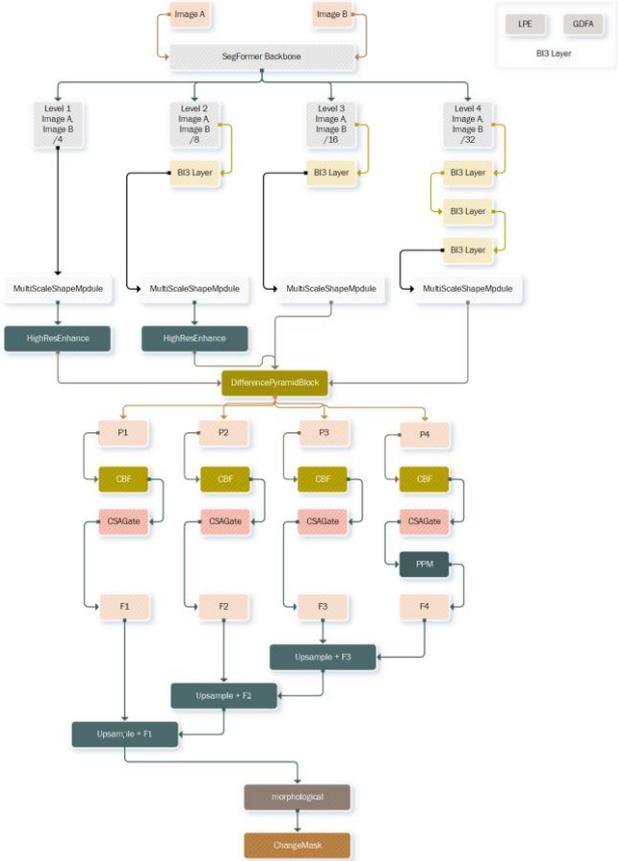

Fig. 1. The overall architecture of the proposed SCA-Net. The pipeline illustrates the Siamese backbone, enhanced BI³ layer, adaptive multi-scale processing (MultiScaleShapeModule & HighResEnhance), Difference Pyramid Block (DPB), attention modules (CSAGate & PPM), and the decoder.

3.5. Adaptive Multi-scale Processing

Recognizing that different feature levels contain distinct information, we apply specialized processing modules adaptively. The MultiScaleShapeModule is applied to the low-resolution, semantically rich features from levels 3 and 4, corresponding to strides of 1/16 and 1/32. This module consists of five parallel branches including three dilated convolutions with rates 1, 2, and 3 to capture extensive context [19], and two directional convolutions with kernels of 1×5 and 5×1 to detect elongated features such as roads [20]. The outputs from all branches are subsequently concatenated and fused. Conversely, the HighResEnhance block processes the high-resolution, detail-oriented features from levels 1 and 2, with strides of 1/4 and 1/8. This block comprises a depthwise separable convolution [21] formed by a depthwise 3×3 convolution followed by a pointwise 1×1 convolution for efficient feature extraction, and is followed by a channel attention layer [15] to adaptively enhance the most crucial feature channels for accurate small object detection.

3.6. Multi-Level Attention for Context and Detail

A Pyramid Pooling Module (PPM) [6] is integrated on the deepest feature map to aggregate multi-scale global context via average pooling at four different bin sizes (1×1, 2×2, 3×3, and 6×6). In parallel, a Channel and Spatial Attention Gate (CSAGate) [7] refines local details by sequentially applying channel attention for re-weighting feature channels and spatial attention for highlighting the most salient locations within the feature map.

3.7. Decoder and Output

The decoder fuses processed information from the PPM, CSAGate, and the refined DPB outputs through a top-down pathway with lateral connections [14]. It uses bilinear upsampling and convolutions to recover spatial resolution, producing a pixel-wise change map with three classes: background, changed road, and changed building.

3.8. Dynamic Training Strategy

Dynamic Composite Loss: We employ a loss function

$$L = \lambda^1 L_C E + \lambda^2 L_D ice + \lambda^3 L_L ovász + \lambda^4 L_C ontrastive \quad (3)$$

The weights $\{\lambda_i\}$ are strategically scheduled across four distinct training phases as outlined in prior work [22]. The initial warm-up phase, covering epochs 0 to 10, focuses primarily on the Cross-Entropy loss L_CE. As training progresses into the refinement phase from epochs 10 to 30, the weights $\lambda_2$ and $\lambda_4$ are increased to promote better shape consistency and feature discrimination. The subsequent optimization phase, spanning epochs 30 to 60, emphasizes the Lovász loss L_Lovász [23] to directly optimize the Intersection over Union metric. Finally, the convergence phase during epochs 60 to 100 employs a balanced combination of all individual loss components.

Layer-wise Learning Rates & Augmentation: A higher learning rate is applied to new decoder/classification head layers for rapid learning, while a lower rate is used for the pre-trained backbone to prevent catastrophic forgetting [24]. Coordinated data augmentation (MixUp [25], CutMix [26]) is applied identically to both images in a pair ($I_1$, $I_2$) and their label mask, preserving the critical temporal relationship.

IV. EXPERIMENTS AND RESULTS

Setup: We use LEVIR-MCI [3] (10,077 pairs, 3 classes) as the primary dataset and LEVIR-CD [27] (637 pairs, binary) for broader comparison. Metrics include Overall Accuracy (OA), mean F1-Score (F1), mIoU, mean Precision, mean Recall, and IoU for Small Buildings (<400 pixels). All models were implemented in PyTorch, evaluated on an NVIDIA GeForce GTX 1650 using AdamW, with Change-Agent [3] as the baseline.

4.1. Quantitative Results

Comparison with Change-Agent [3]: Table I shows SCA-Net's superior performance on both datasets. The most notable result is the 57.9% improvement in Small Building IoU on LEVIR-MCI. The significant boost in Recall, coupled with maintained Precision, indicates a superior ability to detect true changes with fewer false alarms. Our training strategy reduced required epochs from 200 to 78, a 61% reduction in training time, while inference speed remained comparable (15.19 FPS vs. 14.7 FPS).

**Table I: Quantitative Comparison with Change-Agent [3]**

| METRIC | CHANGE-AGENT[3] | SCA-NET (PROPOSED) | IMPROVEMENT |
|---|---|---|---|
| OVERALL ACCURACY | 0.9822 | **0.9838** | +0.16% |
| MEAN F1-SCORE | 0.9258 | **0.9416** | +1.58% |
| MIOU | 0.8654 | **0.8918** | +2.64% |
| MEAN PRECISION | 0.9350 | **0.9409** | +0.59% |
| MEAN RECALL | 0.9170 | **0.9425** | +2.55% |
| SMALL BUILDING IOU | 0.1724 | **0.7514** | +57.90% |

Comparison with State-of-the-Art: Table II compares SCA-Net with other methods on LEVIR-CD. Results for baseline methods are taken from their original publications. SCA-Net achieves the highest reported F1-Score and IoU, establishing its competitive edge.

**Table II: Comparison with State-of-the-Art on LEVIR-CD**

| MODEL / METHOD (YEAR) | F1-SCORE | IOU |
|---|---|---|
| BIT (2021) [28] | 0.8884 | 0.7992 |
| CHANGEFORMER (2022) [13] | 0.8232 | 0.6996 |
| DMINET (2023) [29] | 0.8431 | 0.7288 |
| SNUNET (2021) [12] | 0.8608 | 0.7556 |
| **SCA-NET (PROPOSED)** | **0.9462** | **0.9052** |

Ablation Study: Table III, conducted on LEVIR-MCI, confirms the incremental contribution of each proposed component. The enhanced BI³ layer boosted Recall. The Adaptive Multi-scale Processing module improved IoU for roads. The full integration of all components with the optimized training strategy delivered the best overall performance.

**Table III: Ablation Study on LEVIR-MCI**

| MODEL GROUP | ACTIVE MODULES | MIOU | IOU ROAD | IOU BUILDING |
|---|---|---|---|---|
| CHANGE-AGENT [3] | Base Architecture | 0.8654 | 0.7954 | 0.8200 |
| MODEL 1 | + Enhanced BI³ (LPE+GDFA) | 0.8751 | 0.8188 | 0.8273 |
| MODEL 2 | + Adaptive Multi-scale Processing | 0.8815 | 0.8333 | 0.8307 |
| MODEL 3 | + Difference Pyramid Block | 0.8822 | 0.8359 | 0.8302 |
| MODEL 4 | + Attention (PPM+CSAGate) | 0.8849 | 0.8323 | **0.8410** |
| **FINAL MODEL (SCA-NET)** | All + Optimized Training | **0.8918** | **0.8534** | 0.8399 |

4.2. Qualitative Analysis

Fig. 2 provides compelling visual evidence of SCA-Net's strengths compared to Change-Agent [3] and Ground Truth (GT).

In contrast to the fragmented predictions generated by the Change-Agent model [3], SCA-Net produces more complete and topologically connected road segments, as visually demonstrated in Figure 2, Rows 1 and 2. This significant improvement in continuous road detection is primarily attributed to our novel shape-aware processing, which utilizes directional convolutions to effectively capture elongated linear structures, and our robust multi-scale feature fusion strategy that ensures consistent information flow across different resolution levels, thereby preserving the structural integrity of roads throughout the network.

In challenging scenarios characterized by significant lighting variations and shadows, our model demonstrates greater robustness by generating notably fewer false positives, as evidenced in Figure 2, Rows 3 and 4. This enhanced stability and reduction of false alarms are attributable to the comprehensive global context captured by the Pyramid Pooling Module (PPM) and the robust feature interaction facilitated by our enhanced BI³ layer, which collectively enable the model to more effectively discern real semantic changes from mere transient appearance variations.

The results presented in Figure 2, Rows 5, 6, and 7, clearly demonstrate that SCA-Net successfully identifies and accurately segments small buildings, which are either entirely missed or poorly delineated by the baseline. This compelling visual evidence aligns perfectly with the substantial quantitative improvement and is a direct consequence of the synergistic operation of our novel

Difference Pyramid Block and the CSAGate attention mechanism, which work in tandem to enhance the model's focus on fine-grained details and amplify local contrasts crucial for detecting small objects.

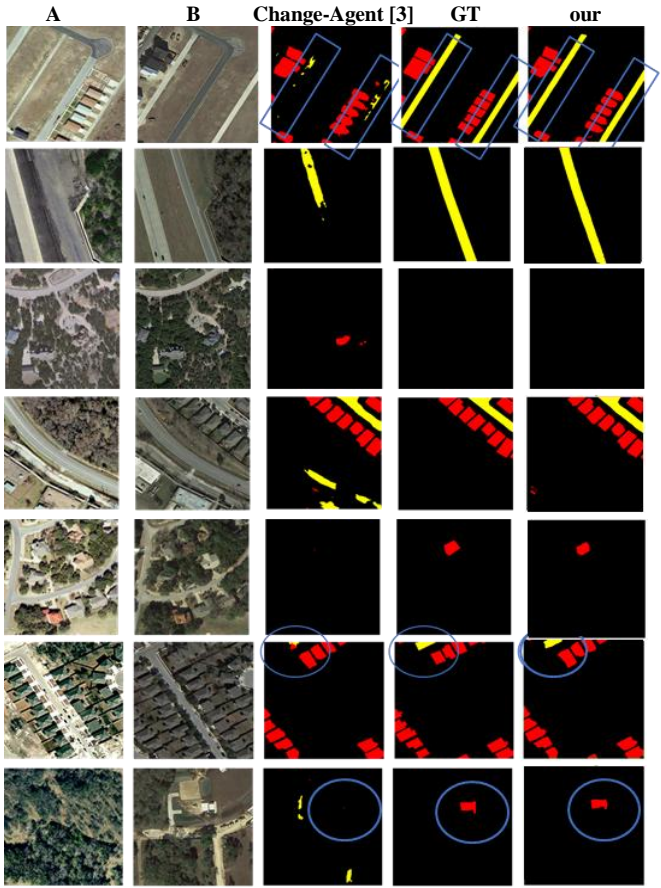

Fig. 2. presents a qualitative comparison of change detection results. From top to bottom, the visual results demonstrate key improvements of the proposed method. In rows 1 and 2, enhanced continuity in road detection is evident as SCA-Net produces more connected and complete road segments compared to the fragmented outputs of Change-Agent [3]. Rows 3 and 4 showcase improved robustness to challenging illumination variations and shadows, where SCA-Net generates significantly fewer false positives. Furthermore, rows 5, 6, and 7 highlight superior detection and segmentation capabilities for small buildings, which are largely missed or poorly delineated by the baseline model. Across all examples, the columns from left to right consistently display Image A from Time T1, Image B from Time T2, the prediction from Change-Agent [3], the Ground Truth (GT) mask, and finally the prediction from the proposed SCA-Net.

## V. Conclusion

This paper introduces **SCA-Net**, a comprehensively enhanced architecture that effectively tackles persistent challenges in remote sensing change detection, such as low sensitivity to small objects. By integrating a novel Difference Pyramid Block, an Adaptive Multi-scale Processing module, and advanced attention mechanisms (PPM and CSAGate), SCA-Net achieves more robust and precise change detection. A novel dynamic training strategy with a composite loss function ensured stable convergence and reduced required training time by 61%. Extensive experiments demonstrate that SCA-Net sets a new state-of-the-art performance, achieving a remarkable 57.9% improvement in IoU for small buildings. A limitation of our current work is its focus on RGB imagery and specific change classes (buildings/roads). Future work will extend SCA-Net to multi-spectral and SAR data, generalize it for multi-class change detection, and explore integration with vision-language models for interactive analysis.